\documentclass[letterpaper]{article}
\usepackage{comment}
\usepackage{ijcai22}
\usepackage{times}
\usepackage{helvet}
\usepackage{courier}
\usepackage{amsmath,pgfmath,pgffor}
\usepackage{mathtools,amssymb}
\usepackage{graphicx}
\usepackage{booktabs}
\usepackage[table]{xcolor}
\usepackage{color, colortbl}
\usepackage{afterpage}
\usepackage{times}
\usepackage{soul}
\usepackage{url}
\usepackage[hidelinks]{hyperref}
\usepackage[utf8]{inputenc}
\usepackage[small]{caption}
\definecolor{LightCyan}{rgb}{0.92,1.0,0.97}
\newcolumntype{a}{>{\columncolor{LightCyan}}c}
\newcommand\blfootnote[1]{%
  \begingroup
  \renewcommand\thefootnote{}\footnotetext{#1}%
  \endgroup
}

\begin{document}

\title{Deciphering Environmental Air Pollution with Large Scale City Data}

\author{
Mayukh Bhattacharyya$^{*1}$
\and
Sayan Nag$^{*2}$\And
Udita Ghosh$^{3}$
\affiliations
$^1$Stony Brook University\\
$^2$University of Toronto\\
$^3$Zendrive Inc.\\
\emails
mayukh.bhattacharyya@stonybrook.edu,
sayan.nag@mail.utoronto.ca,
uditag@zendrive.com
}

\maketitle
\begin{abstract}
Air pollution poses a serious threat to sustainable environmental conditions in the 21st century. Its importance in determining the health and living standards in urban settings is only expected to increase with time. Various factors ranging from artificial emissions to natural phenomena are known to be primary causal agents or influencers behind rising air pollution levels. However, the lack of large scale data involving the major artificial and natural factors has hindered the research on the causes and relations governing the variability of the different air pollutants. Through this work, we introduce a large scale city-wise dataset for exploring the relationships among these agents over a long period of time. We also introduce a transformer based model - cosSquareFormer, for the problem of pollutant level estimation and forecasting. Our model outperforms most of the benchmark models for this task. We also analyze and explore the dataset through our model and other methodologies to bring out important inferences which enable us to understand the dynamics of the causal agents at a deeper level. Through our paper, we seek to provide a great set of foundations for further research into this domain that will demand critical attention of ours in the near future.
\end{abstract}

\section{Introduction}

\def\thefootnote{*}\footnotetext{denotes equal contribution}\def\thefootnote{\arabic{footnote}}

\def\thefootnote{$^1$}\footnotetext{Dataset and Code: https://github.com/mayukh18/DEAP}

Advancement of civilization has led to a lot of interferences on earth generated by humans. While a lot of those pose a threat to the balance of ecosystem and overall climate of the planet, air pollution happens to hold severe and immediate impacts on us humans. PM2.5 and NO$_2$ the two most common air pollutants are well known to inflict irreversible respiratory disease\cite{lung1}. Besides asthma attacks and cardiovascular issues, it has been observed to cause or exacerbate cancers, diabetes \cite{diabetes} and also influence mortality in infants \cite{infants}. The major sources of pollutants like PM2.5, NO$_2$, O$_3$ etc are emissions from automobiles, power plants and other heavy industries$^2$ \afterpage{\blfootnote{$^2$NO$_2$:https://www.epa.gov/\{no2-pollution, pm-pollution\}}}. The close proximity of industrial zones around highly populated metropolitan areas combine all the sources of the pollutants to create a very poor living conditions in quite a few cities in the world. Governments of multiple cities have tried methodologies ranging from artificial rains by cloud-seeding, partial traffic ban to giant air purifiers. All these efforts showcase the rising importance of the issue with every passing year. A year long stretch of lockdown and work-from-home systems has suddenly proved how the absence of public human activities has an improving effect on the pollution levels in big metropolitan cities. Traffic, a big factor behind urban air pollution was completely absent in the initial 1-2 months. This led to significant drop in different pollutant levels as demonstrated in different studies such as \cite{covid1}. These developments have garnered a lot of attention over finding viable solutions that was inconceivable on a large scale earlier due to lack of data with such variability. We feel this is an exciting opportunity to hasten the research in this domain further. In this regard we present our dataset and methodology which we believe will aid in our common mission.

The primary contributions of the paper are three-fold:

\begin{enumerate}

\item We have introduced a large-scale curated spatio-temporal dataset$^1$ encompassing daily levels of different pollutants and the major casual agents (both artificial and natural) for a duration over 2 years spanning over more than 50 cities in the United States. As per our knowledge, this is the largest dataset in terms of it's coverage of the number of cities. Also, alongside traffic emissions, we adopted a novel method to quantify the critical impact of emissions from power plants, which is a first for studies on urban air pollution till date. 

\item We have proposed a non-linear re-weighting attention mechanism for transformers which weights neighboring tokens more with respect to the far-away ones enforcing a strict locality constraint. Although, similar weighting schemes have been very recently adopted for language modeling tasks but this is the first time such a method is being used for time-series analysis. Furthermore, to the best of our knowledge this is the first attempt to use transformers for multi-variate pollutant forecasting tasks. In addition, we have considered a novel hybrid loss function combining Mean Squared Error and Dynamic Time Warping resulting in more robust similarity computation between two temporal sequences. 

\item We have also presented a holistic analysis of the dataset that we are introducing. With bayesian modeling, we have captured the relative importances of the different factors in influencing the pollutant levels. We have also analysed the dependency of the pollutants on previous days values thus reflecting the duration of retention of pollutants in the atmosphere. Additionally, we have presented different inferences drawn from the data which brings out actionable information that can used on a wider scale.

\begin{figure*}[h]
    \centering
    \includegraphics[width = 0.3\textwidth,height = 0.23\textwidth]{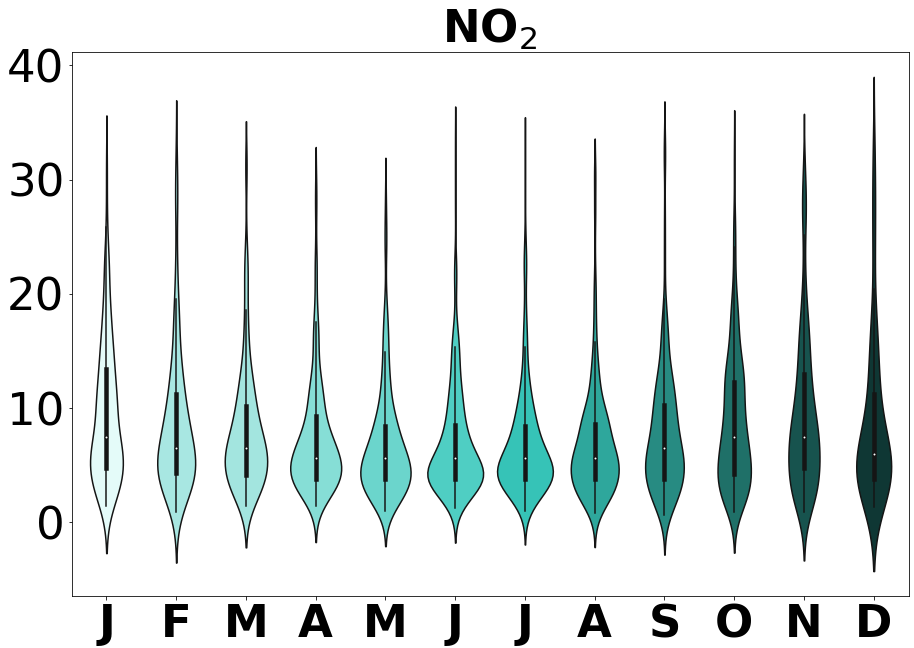}
    \includegraphics[width = 0.3\textwidth,height = 0.23\textwidth]{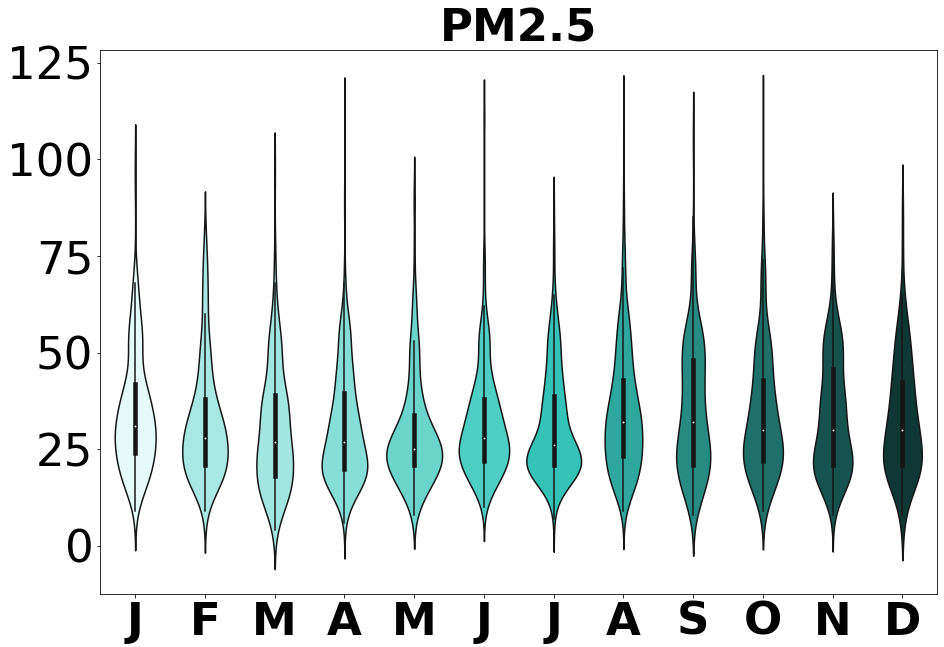}
    \includegraphics[width = 0.3\textwidth,height = 0.23\textwidth]{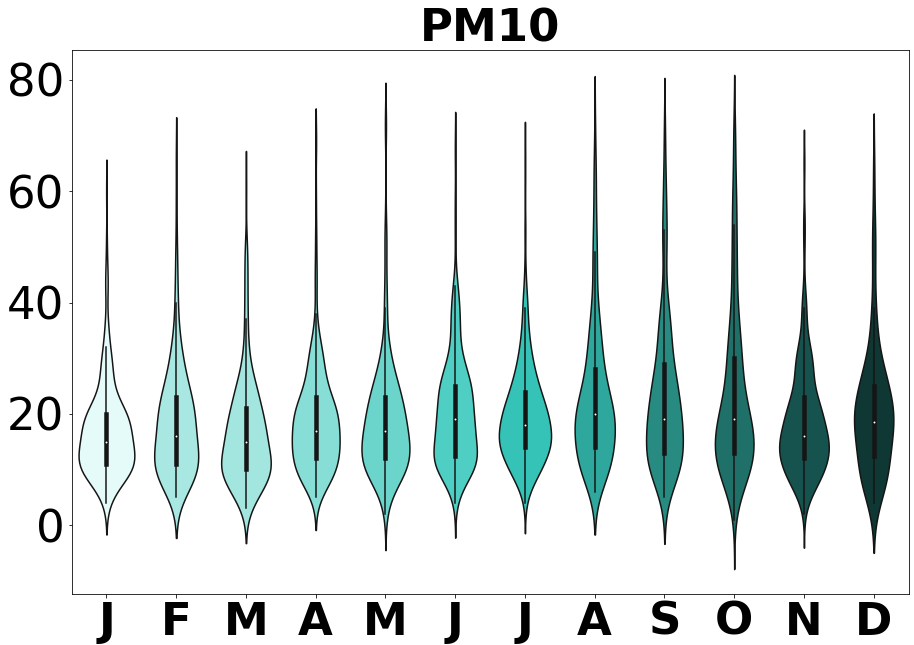}
    \includegraphics[width = 0.3\textwidth,height = 0.23\textwidth]{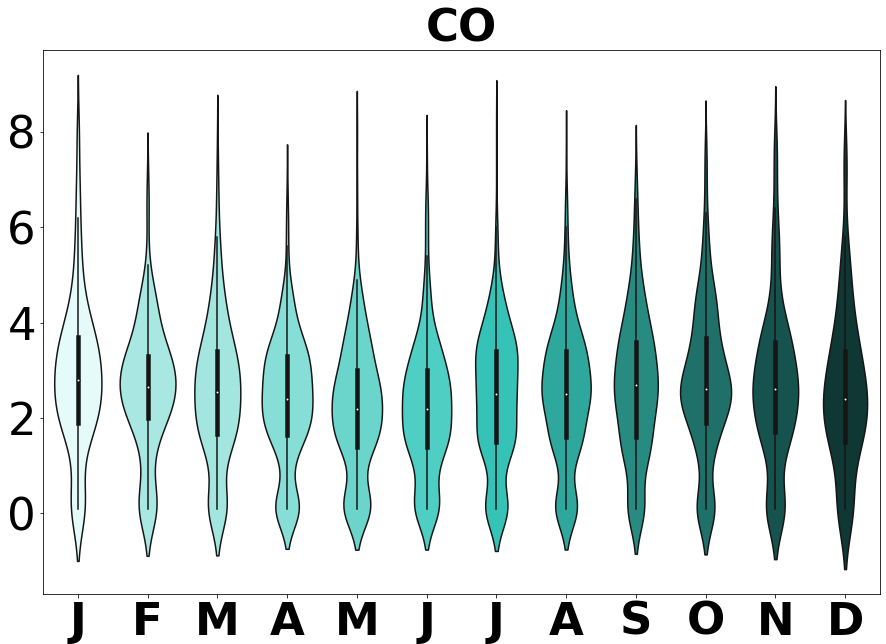}
    \includegraphics[width = 0.3\textwidth,height = 0.23\textwidth]{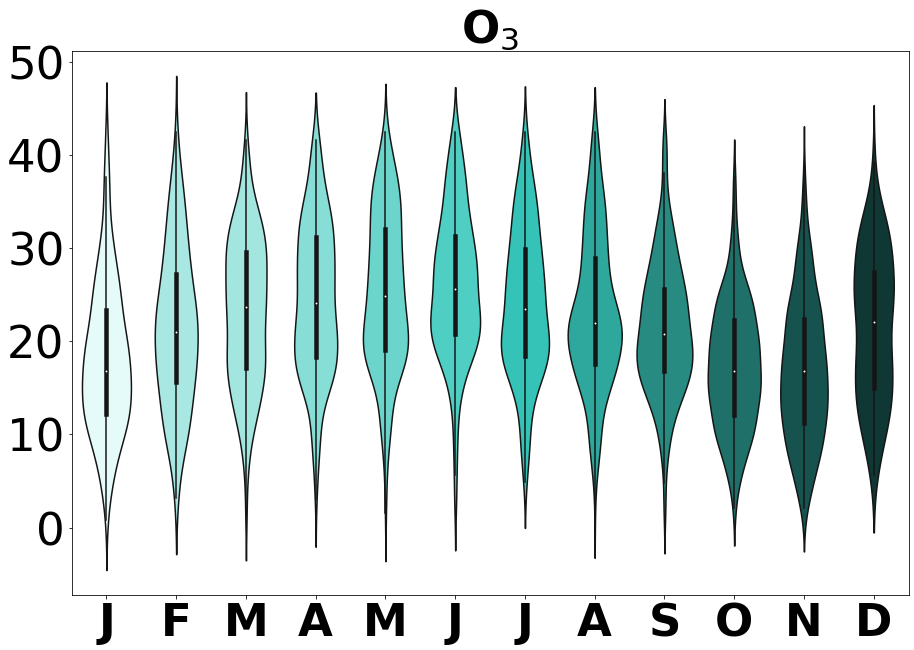}
    \includegraphics[width = 0.3\textwidth,height = 0.23\textwidth]{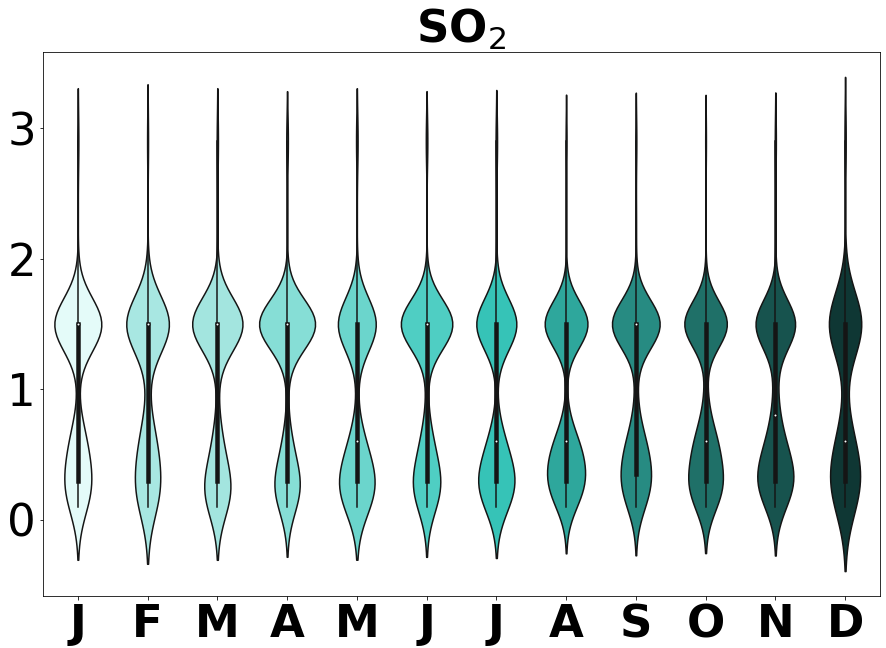}
    \caption{Distribution of NO$_{2}$, PM2.5, PM10, CO, O$_3$, and SO$_2$ over a period of 12 months (x-axis). Each of the pollutants demonstrate a different distribution both over the period of 12 months and also among the pollutants on each month.}%
    \label{fig:violin_plots}
\end{figure*}

\end{enumerate}
\section{Related Work}

Study of factors leading to air pollution has gained momentum in recent time, although it is a persistent problem for long. Although the studies have taken different problem statements but they have focused mainly around PM2.5 as the central theme. Though previous works existed for air quality forecasting, one of the first works to consider natural influencers like wind, humidity, temperature as well as gases like NO, CO was done by \cite{russo}. Traditional machine learning methods like Support Vector Machines  have been used to forecast Air Quality Index (AQI) as well as individual pollutant levels in the air \cite{aqical}. Prediction of specific pollutants concentration like PM2.5 by gradient boosting approach from past data of PM2.5 concentration and climate information is presented in \cite{aqi_taiwan}. However, with the advent of RNNs \cite{rumelhart1985learning,sherstinsky2020fundamentals}, most of the recent works have been using LSTMs \cite{lstm} for air pollution estimation \cite{qadeer2020long}. \cite{li} and \cite{lstmconc} utilised LSTM based systems to predict the concentration of the air pollutants. A Spatio-temporal DNN has been presented in \cite{soh} which takes surrounding conditions into consideration while predicting. Leveraging the information from air quality monitoring stations and other factors like city's points of interests, road networks and meteorological data, a combination of feed-forward and recurrent networks has been used to model static and sequential data. \cite{cheng2018neural}. Considering air quality data,
meteorology data and weather forecast data, a deep distributed fusion network has been used with a spatial transformation component for predicting air quality of respective monitoring stations \cite{yi2018deep}. Using multi-level attention networks with spatial-temporal and meteorological data, air and water quality prediction have been done in \cite{liang2018geoman}.

However, the drawbacks of most of these works is either they are concentrated on a single region which makes the models not universal, or that they do not consider the influences of causal agents of pollution like automobile and industry emissions. Although few studies like \cite{thermalpp} evaluates the effectiveness of several thermal power plant control measures on the air quality, a larger exploration or forecasting study is not available due to lack of large scale data.


\section{Dataset}
\label{headings}

We present a large dataset for modeling the variation of air pollution at the daily level over multiple cities involving data from most of the influencing agents both natural and man-made. The dataset is the largest as per our knowledge in regards to the number of locations and days involved. 

Overall, the dataset contains a total of 35,596 unique sample points spanning 54 cities and 24 months with each sample point representing a unique (date, city) combination. The data is collected and curated from multiple sources. Hence, some cities and some dates do not have the values for all the pollutants and features. The different aspects of the dataset are given in Table 1. The sources of the features and the data processing involved are described in the respective sections below:

\begin{table}[htp]
    \footnotesize
    \centering


    
    \begin{tabular}{ccc}

    \toprule
    \textbf{Pollutants} & \textbf{Valid Samples} & \textbf{Valid Cities}\\[0.5ex]
    \midrule
    PM2.5 & 35134 & 54\\
    \rowcolor{LightCyan}
    PM10 & 16965 & 29\\
    O3 & 33950 & 54\\
    \rowcolor{LightCyan}
    SO2 & 14676 & 39\\
    NO2 & 23558 & 41\\
    \rowcolor{LightCyan}
    CO & 24538 & 42\\
    \bottomrule
    \end{tabular}\\[0.5ex]
    
    \caption{Dataset Statistics. A city is considered valid here if it has at least 2 months data of the pollutant levels.}
    \label{tab:my_label}
\end{table}


\begin{itemize}

\item \textbf{Air Pollutants} The daily data of different pollutant species at a city level was obtained from Air Quality Open Data Platform$^3$\afterpage{\blfootnote{$^3$https://aqicn.org/data-platform/covid19}}. The min, max and median values of the pollutant on a day are provided. The concentration values are normalized in US EPA standard. There are six air pollutants in our dataset namely NO$_2$, PM2.5, PM10, SO$_2$, O$_3$, and CO. The violin plots in Figure \ref{fig:violin_plots} illustrates the monthly distribution and variation of the aforementioned air pollutants.

\begin{figure}[t!]
    \centering
    \includegraphics[width=0.4\textwidth]{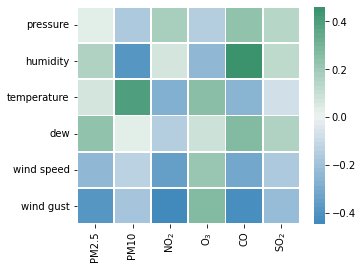}
    \caption{Correlation of Meteorological Factors with Pollutant levels}%
    \label{fig:natural_factors_corr}
\end{figure}

\item \textbf{Meteorological Factors} Meteorological factors like humidity, windspeed, temperature and pressure have an impact on the concentration of pollutants in the atmosphere. The concentration values of these meteorological factors are also obtained from the Air Quality Open Data Platform$^3$ like above. They serve as input features for our models. The units of the features are provided in the dataset. Figure \ref{fig:natural_factors_corr} depicts the correlations among these meteorological factors and the respective pollutants.

\item \textbf{Traffic} The corresponding daily traffic data is collected from$^4$\afterpage{\blfootnote{$^4$https://data.bts.gov/Research-and-Statistics/Trips-by-Distance/w96p-f2qv}} provided by Maryland Transport Institute \cite{maryland}. The traffic data follows almost the same spatio-temporal granularity as the air pollutant data, apart from one aspect. The traffic data is provided at a county level, not at the city level. But since we are dealing with mainly major metropolitan areas, we have taken the liberty to consider the traffic of the city same as the county it lies within. The trip-based data from \cite{maryland} is processed to collate all the trips in a day to calculate the "million miles" of travel in the day, which we treat as the measure of traffic in that city for that day.

\item \textbf{Power Plant Emission} The data around the generation patterns of power plants could only be obtained at a monthly level from US EIA Website$^5$\afterpage{\blfootnote{$^5$https://www.eia.gov/electricity/data/eia923/}}. Considering the production patterns of power plants don't change much at the daily level, we made a pragmatic approximation of averaging the monthly value to the daily level. There are 11,833 power plants we have considered in the dataset. It should however be noted that we have only selected generation data of generators running on fuel types - Coal, Oil, Gas and Biomass, since these are the major ones most frequently held responsible for air pollution.

While we do provide the power-plants data in it's granular raw form, we needed a single feature representing the effects of the power plants for a certain (city,date) pair. For that purpose, we design an intuitive metric to form the feature .

\begin{equation}
    {I_{pp}}_{c,t} = \sum_p G_{p} / r_{cp}^2, \;\;\; \textit{for}\:\, r_{cp} < R_{limit}
\end{equation}

where $I_{pp}$ is the feature obtained from power-plants for a city $c$ on a date $t$. $G_{p}$ is the average daily generating capacity for the plant for that month and $r_{cp}$ is the linear distance between the power-plant and the centre of the city. We have taken $R_{limit}$ as 30 km.

\end{itemize}

\begin{table}[!h]\footnotesize
    \centering
    \begin{tabular}{cccc}

    \toprule
    \textbf{Features} & \textbf{25\%} & \textbf{50\%} & \textbf{75\%} \\[0.5ex]
    \midrule
    Pollutants (X) & & & \\[0.5ex]
    \rowcolor{LightCyan}
    1. PM2.5 & 21 & 28 & 39 \\[0.5ex]
    2. PM10 & 10 & 15 & 21 \\[0.5ex]
    \rowcolor{LightCyan}
    3. NO2 & 3.8 & 6.4 & 10.2 \\[0.5ex]
    4. O3 & 14.5 & 20.8 & 27.2 \\[0.5ex]
    5. SO2 & 0.3 & 1.1 & 1.5 \\[0.5ex]
    \rowcolor{LightCyan}
    6. CO & 1.8 & 2.5 & 3.6 \\[0.5ex]
    Traffic Distance(I$_T$) & 19.54 & 31.32 & 49.44 \\[0.5ex]
    \rowcolor{LightCyan}
    Power Plant Emission(I$_{pp}$) & 0.40 & 1.69 & 6.95 \\[0.5ex]
    \bottomrule
    \end{tabular}\\[0.5ex]
    \caption{Feature Distributions. Since the data is spread over 2 years in 54 cities, we get a good distribution of the feature values.}
    \label{tab:feat_distribution}
\end{table}


\begin{table*}[!htp]\footnotesize
    \centering
    \begin{tabular}{l cccccc cccccc}

    \toprule
    
    \textbf{Method} & \multicolumn{6}{c}{\textbf{RMSE}} & \multicolumn{6}{c}{\textbf{MAPE (\%)}}\\[0.5ex]
    \cmidrule(lr){2-7}
    \cmidrule(lr){8-13}
    
     & \textbf{PM2.5} & \textbf{PM10} & \textbf{NO$_{2}$} & \textbf{O$_{3}$} & \textbf{CO} & \textbf{SO$_{2}$} & \textbf{PM2.5} & \textbf{PM10} & \textbf{NO$_{2}$} & \textbf{O$_{3}$} & \textbf{CO} & \textbf{SO$_{2}$}\\[0.5ex]
    \midrule
    OLS & 14.06 & 9.63 & 4.34 & 8.62 & 5.78 & 1.95 &
          48.6 & 39.3 & 67.1 & 206.6 & 214.8 & 182.0 \\[.5ex]
    BR & 14.34 & 8.96 & 5.69 & 16.11 & 6.88 & 1.95 & 
                43.2 & 63.5 & 88.9 & 378.0 & 458.8 & 223.3\\[.5ex]
    GBM & 12.78 & 10.14 & 3.60 & \textbf{6.94} & 5.44 & 1.94 & 36.1 & \textbf{38.2} & 46.3 & 181.8 & \textbf{71.7} & 133.5\\[.5ex]
    LSTM & 12.61 & 8.44 & 3.60 & 8.05 & 5.53 & 1.76 &
                42.6 & 52.9 & 54.9 & 174.6 & 170.0 & 95.2 \\[.5ex]
    LSTM E & 13.43 & 7.85 & 4.10 & 8.02 & 5.50 & 1.87 &
                43.5 & 45.9 & 63.1 & 179.5 & 203.8 & 143.3\\[.5ex]
    Transformer & 11.89 & 8.08 & 3.59 & 8.17 & 5.44 & \textbf{1.72} &
                36.1 & 43.6 & 48.8 & 152.6 & 157.9 & 73.0\\[.5ex]
    cosFormer & 11.88 & 8.10 & 3.59 & 8.19 & \textbf{5.42} & 1.76 &
                35.8 & 45.2 & 48.5 & 156.1 & 138.8 & 78.1\\[.5ex]
    \rowcolor{LightCyan}
    \textbf{\texttt{cosSquareFormer}} & \textbf{11.68} & \textbf{8.06} & \textbf{3.49} & 8.14 & \textbf{5.42} & 1.75 &
                \textbf{34.7} & 45.9 & \textbf{43.5} & \textbf{146.6} & 125.4 & \textbf{69.1}\\[.5ex]
    
    \bottomrule
    \end{tabular}\\[0.5ex]
    \caption{Performance of predictions from different models for all 6 pollutants. LSTM E and Attention LSTM E are trained on explicit information of weekday and month whereas the explicit information have been excluded whilen training the remaining models. The sequence length (number of past days) for all the LSTM and Transformer (including variants) is 7 days.}
    \label{tab:results}
\end{table*}

\section{Proposed Method}

The general form of transformer is given as: $T(x) = F(A(x) + x)$ where $T$ is the transformer block with an input sequence $x$, $F$ is the feed-forward network and the self-attention function is given by $A$ which has a quadratic space and time complexity depending on the length of the input sequence ($N$). Attention function has three learn-able linear matrices namely Query ($Q$), Key ($K$) and Value ($V$) \cite{vaswani2017attention} which when combined together using a dot-product attention with softmax normalization gives the final attention output given as:

\begin{equation}
    O_i = \sum_j\frac{e^{Q_iK_j^T}}{\sum_je^{Q_iK_j^T}}V_j, \; \forall i,j \in N
\end{equation}

The quadratic time and space complexity poses a computational challenge, especially for long sequences. Few solutions, including linearization of self-attention were proposed in this regard \cite{choromanski2020rethinking,katharopoulos2020transformers}. Along the same lines, a recently proposed work \cite{anonymous2022cosformer}, used a linear operation with decomposable non-linear cosine-based re-weighting mechanism instead of a standard non-linear softmax operation. They noticed that non-linear re-weighting introduced by softmax attention results in a stable training process and addition of a decomposable cos-based re-weighting scheme can introduce recency bias to the attention matrix. As a consequence, locality is enforced.

Inspired by the above idea, in order to enforce stricter locality constraint, we proposed a decomposable cosine-square re-weighting mechanism which weights the neighbouring tokens more (compared to cosine) with respect to the far-away ones (see Appendix for Algorithm). This cosine-square re-weighting can also be loosely considered as a linear combination of cosFormer \cite{anonymous2022cosformer} and Linear Transformer \cite{katharopoulos2020transformers}. The similarity function between $Q$ and $K$ with cosine-square re-weighting is defined as:

\begin{equation}
    \begin{split}
        s(\tilde{Q}_i,\tilde{K}_j) = \tilde{Q}_i\tilde{K}_j^Tcos^2\bigg(\pi\frac{i-j}{2M}\bigg)\\
        = \frac{1}{2}\Bigg[\tilde{Q}_i\tilde{K}_j^T + \tilde{Q}_i\tilde{K}_j^Tcos\bigg(\pi\frac{i-j}{M}\bigg)\Bigg]
    \end{split}
\end{equation}

Using Ptolemy's theorem and decomposing the above expression further leads to:

\begin{equation}
\centering
    \begin{split}
        s(\tilde{Q}_i,\tilde{K}_j) = \frac{1}{2}\Bigg[\tilde{Q}_i\tilde{K}_j^T +\\ \bigg(\tilde{Q}_icos\bigg(\pi\frac{i}{M}\bigg)\bigg)\bigg(\tilde{K}_jcos\bigg(\pi\frac{j}{M}\bigg)\bigg)^T +\\ \bigg(\tilde{Q}_isin\bigg(\pi\frac{i}{M}\bigg)\bigg)\bigg(\tilde{K}_jsin\bigg(\pi\frac{j}{M}\bigg)\bigg)^T \Bigg]
    \end{split}
\end{equation}

where $i,j = 1,..., N,\; M \geq N, \; \tilde{Q} = f(Q), \; \tilde{K} = f(K), \; f = ReLU$ and output is given as:

\begin{equation}
    \begin{split}
        O_i = \frac{\sum_{j=1}^Ns(\tilde{Q}_i,\tilde{K}_j)V_j}{\sum_{j=1}^Ns(\tilde{Q}_i,\tilde{K}_j)}
    \end{split}
\end{equation}

\subsection{Loss Function}

For training purposes we have used a novel hybrid loss function composing a weighted combination of MSE Loss and a soft-Dynamic Time Warping (sDTW) loss \cite{cuturi2017soft}. For forecasting purposes, it has been noticed in \cite{cuturi2017soft} that sDTW loss performs superior to standard Euclidean loss because of the former's robustness to similarity computation between two temporal sequences. In our work we have seen combining these two losses together gives better performances and training stability as compared to their individual counterparts. Hence, our proposed loss function between ground-truth ($y$) and predicted ($\Bar{y}$) time-series is given as:

\begin{equation}
    L(y,\Bar{y}) = MSE(y,\Bar{y}) + \lambda \; sDTW(y,\Bar{y})
\end{equation}

We have considered $\lambda = 0.5$ for the experiments.

\section{Experiments}

In this section, we approach the problem of estimating pollutant levels based on information about the causal and influencing factors.

We have evaluated our sequential proposed method with both non-sequential and sequential methodologies as baselines. For non-sequential models, the problem statement is that of estimation: trying to estimate a pollutant value based on the day's features. However, for sequential models, we incorporated the forecasting problem. This means we are trying to predict a certain day's pollutant levels based on features of that day along with the pollutant level of the previous day.

In total, we compared our method with 9 baselines. Among the estimator models, we use Ordinary Least Squares (OLS), Bayesian Regression (BR), Gradient Boosting Machines (GBM). Among the sequential forecasting models, we use LSTM, Attention LSTM, Transformer and cosFormer.

\subsection{Test Data} Since we have both sequential and non-sequential models, we needed an train-evaluation split of the whole dataset which would have let us evaluate sequential models with the same ease as non-sequential traditional models.
Usually for time-series data, there is the prevalent norm of selecting a later portion as the evaluation dataset. However, we realized that in doing so we would be restricting the evaluation to a particular season with not much daily variations caused by the features. Since we have both the year 2019 and 2020 in the dataset, we constituted the evaluation or test dataset by taking a continuous 60-day segment for each city starting from the first week of March 2020. Since this time period marked the onset of the COVID lockdowns, we would get a much better variability in terms of the features and pollutants. Considering some values in the test set might be missing due to reasons we discussed before, the evaluation performance is calculated only on the available and valid test data samples.

\subsection{Metrics}
We evaluate and compare all our methods with 2 metrics: Root Mean Square Error (RMSE) and Mean Absolute Percentage Error (MAPE) \cite{botchkarev2018performance}. A combination of these two will give us a holistic picture of the performance of the models being evaluated. 

The results from our method as well as other sequential and non-sequential models are presented in Table \ref{tab:results}.

\begin{figure}[!t]
    \centering
    \includegraphics[height=0.39\textwidth,width=0.48\textwidth]{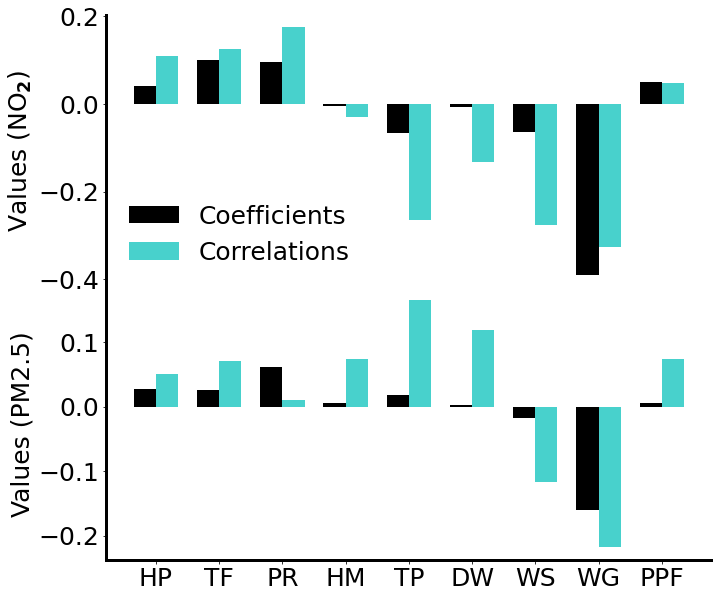}
    \caption{Weights $W_{i}$s from different inputs in BR model alongside associated correlations of these inputs with NO$_{2}$ and PM2.5 levels. X-axis represents (left to right): Population at Home, Traffic, Pressure, Humidity, Temperature, Dew, Wind Gust, Wind Speed and Power Plant Feature.}%
    \label{fig:corr_coeff}%
\end{figure}

%
%

\section{Analysis}

In this section, we explore the findings from our experiments in a little more depth to infer conclusions about the various interrelationships of the features.

\begin{figure}[!h]
    \centering
    {{\includegraphics[width=0.4
    \textwidth, height = 0.28
    \textwidth]{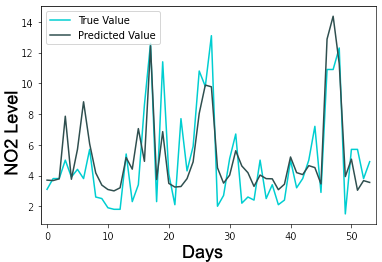} }}
    
    {{\includegraphics[width=0.4
    \textwidth, height = 0.28
    \textwidth]{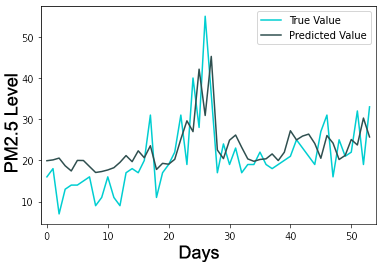} }}
    \caption{General fit of the proposed model on the test set for the city of Las Vegas.}
    \label{fig:las_vegas}
\end{figure}

Table \ref{tab:results} provides a good idea about the general fit and importance of the models in terms of estimating and forecasting pollutant levels. As we can see our proposed \textbf{\texttt{cosSquareFormer}} as well as Gradient Boosting Machines (GBM) do well in terms of performance. As we can see in Figure \ref{fig:las_vegas} our proposed model does a great job in following sudden daily fluctuations in the pollutant levels. 

\begin{figure}[!t]
    \centering
    \includegraphics[width=0.48\textwidth]{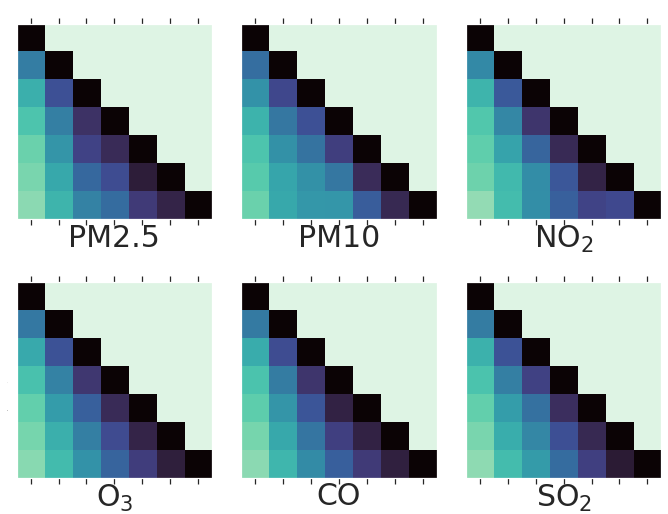}
    \caption{Attention Matrices (mean of all the heads) considering a 7-day forecast period for the third layer of the proposed model for the respective air pollutants.}%
    \label{fig:transformer_attentions}%
\end{figure}

\begin{figure}[!h]
    \centering
    \includegraphics[height=0.27\textwidth,width=0.38\textwidth]{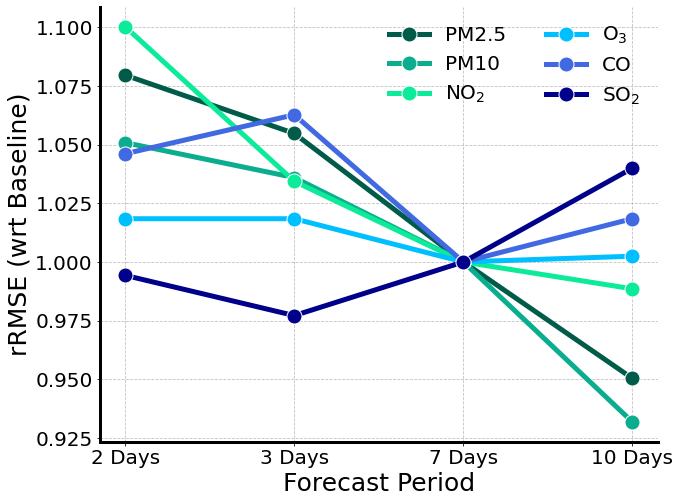}
    \caption{Relative RMSE (rRMSE = RMSE/RMSE$_{Baseline}$) scores for different sequence lengths with respect to the 7-day baseline for all the pollutants with \textbf{\texttt{cosSquareFormer}}.
    }
    \label{fig:cossqformerablations}%
\end{figure}

The good performance of GBM for certain pollutants does raise a question of whether features of past days influence the pollutant levels of future. In order to explore the questions related to the sequential nature of pollutants, we designed an ablation study with multiple sequence lengths with the same experimental setup to maintain parity for modeling. The results given in Figure \ref{fig:cossqformerablations} show that pollutants like PM2.5, PM10 and NO$_2$ have a better performance with longer sequence lengths, whereas the others either degrade or show a flat trend. Thus it can be assumed that the daily concentration of some pollutants indeed have a good dependence on past concentrations whereas some others are mostly independent of it. This can also be analysed at depth from the attention maps of \textbf{\texttt{cosSquareFormer}} in Figure \ref{fig:transformer_attentions}. The values on the last rows(row 6) denote the dependency of pollutant levels on a particular day on the features of the previous 6 days.

    

We also wanted to model the uncertainty in the data through Bayesian Inference. Figure \ref{fig:corr_coeff} shows the weights (means, $\mu$) obtained as a result of Bayesian Inference for NO$_{2}$ and PM2.5 alongside correlation values computed corresponding to the pollutants. This plot not only gives us an idea of the importance of each factor and the extent of the influences of each input feature on affecting the pollutant levels, but also demonstrates a parity that exists between the weights and the corresponding correlation values. The complex nature of Transformers in computing the weights for each feature made it difficult to extract the importance of features from it which we would have shed more light on the significance of features.


\begin{figure}[!h]
    \centering
    {{\includegraphics[width=0.48
    \textwidth, height = 0.30
    \textwidth]{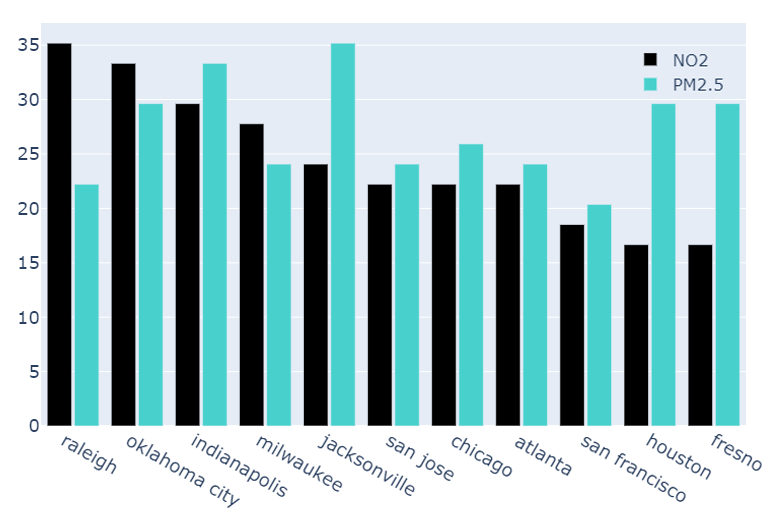} }}
    \caption{Cities with pollutant levels significantly higher than that predicted by the universal model. The y-axis denotes the \% of samples of the city with original pollutant value higher than 125\% of the predicted value.}
    \label{fig:outliers}
\end{figure}

The visualizations shown in Figure \ref{fig:outliers} provide some information about each city's conformity with the universal model. It shows us the cities which have pollutant levels which were much higher than that estimated by our model. It provides us the leads to explore the context and reason behind each such outlier city. An analysis on this basis will provide researchers to identify problematic cases in a meaningful way instead of just flagging cities with high pollutant levels.

\section{Conclusions and Future Works}

Air pollution will lead to be one of the crucial issues of the society in the years to come. An early initiative to tackle the problem may make a big difference in the future. Through our dataset and methodology we have intended to establish a foundation for the community to build on. Our dataset captures a variety of factors influencing the air pollution levels. In this study, we have illustrated the impact of such factors on the air quality indices using extensive studies exploring the various relationships governing the pollutant levels.

Our intention is to improve and extend the dataset with more data considering other emission sources. We believe there are also scopes of further studies like spatio-temporal analysis and other explorations on this dataset itself that may uncover valuable inferences which may progress our understanding of this domain further.

\bibliographystyle{named}
\fontsize{9.0pt}{10.0pt}
\bibliography{ref}

\end{document}


\section{Appendix}

\subsection{Pseudocode for cosSquareFormer}

\begin{minipage}{0.88\textwidth}
\begin{algorithm}[H]
    \centering
    \caption{cosSquareFormer Attention}\label{algorithm}
    \begin{algorithmic}
        \State \text{}
        \State \textbf{Input} \text{$Q \in \mathbb{R}^{N \times d_1}$, $K \in \mathbb{R}^{M \times d_2}$, $V \in \mathbb{R}^{M \times d_2}$}
        \State \textbf{Output} \text{$O \in \mathbb{R}^{N \times d_2}$}
        \State \textbf{Initialize} \text{$A[i] \gets \frac{\pi i}{2N}, \; O[i][j] \gets 0, \; \forall i = 1, ..., N, \; \forall j = 1, ..., d_2$}
        \State \textbf{Initialize} \text{$S^{cos}[i][j] \gets 0, S^{sin}[i][j] \gets 0, T^{cos}[i] \gets 0, T^{sin}[i] \gets 0, S^{only}[i][j] \gets 0, T^{only}[i] \gets 0$}
        \State \text{}
        \For{$i\gets 1, M$}
        \State $\text{$K^{cos}_{i} \gets K_{i}cos\left(\frac{\pi i}{M}\right)$}$ 
        \State $\text{$K^{sin}_{i} \gets K_{i}sin\left(\frac{\pi i}{M}\right)$}$ 
        \State $\text{$S^{cos} \gets S^{cos} + (K^{cos}_{i})^{T}V_{i}$}$
        \State $\text{$S^{sin} \gets S^{sin} + (K^{sin}_{i})^{T}V_{i}$}$
        \State $\text{$S^{only} \gets S^{only} + (K_{i})^{T}V_{i}$}$
        \State $\text{$T^{cos} \gets T^{cos} + K^{cos}_{i}$}$
        \State $\text{$T^{sin} \gets T^{sin} + K^{sin}_{i}$}$
        \State $\text{$T^{only} \gets T^{only} + K_{i}$}$
        \EndFor
        \State \text{}
        \For{$i\gets 1, N$}
        \State $\text{$Q^{cos}_{i} \gets Q_{i}cos\left(\frac{\pi i}{M}\right)$}$ 
        \State $\text{$Q^{sin}_{i} \gets Q_{i}sin\left(\frac{\pi i}{M}\right)$}$ 
        \State $\text{$O_{i} \gets \frac{Q_{i}S^{only} + Q^{cos}_{i}S^{cos} + Q^{sin}_{i}S^{sin}}{Q_{i}T^{only} + Q^{cos}_{i}T^{cos} + Q^{sin}_{i}T^{sin}}$}$
        \EndFor
        \State \text{}
    \end{algorithmic}
\end{algorithm}
\end{minipage}